\documentclass[twoside,11pt]{article}

\usepackage{jmlr2e}
\usepackage{array}
\usepackage{amsmath,amssymb}

\usepackage{booktabs}

\usepackage{subcaption}
\usepackage{mathabx}

\firstpageno{1}

\begin{document}

\title{Modeling Polypharmacy and Predicting Drug-Drug Interactions using Deep Generative Models on Multimodal Graphs}

\author{\name Nhat Khang Ngo $^*$ \email khangnn3@fsoft.com.vn \\
       \addr FPT Software AI Center\\
        Hanoi, Vietnam\\
       \AND
       \name Truong Son Hy $^*$ $^\dagger$ \email tshy@ucsd.edu \\
       \addr Halıcıoğlu Data Science Institute \\ University of California San Diego \\
       La Jolla, CA 92093, United States \\
       \AND 
       Risi Kondor \email risi@uchicago.edu \\
       \addr  Department of Computer Science \\ 
       University of Chicago \\
       Chicago, IL 60637, United States}
\maketitle 

\thanks{$*$: Co-first authors. $\dagger$: Correspondent author.}

\begin{abstract}
Latent representations of drugs and their targets produced by contemporary graph autoencoder models have proved useful in predicting many types of node-pair interactions on large networks, including drug-drug, drug-target, and target-target interactions. However, most existing approaches model either the node's latent spaces in which node distributions are rigid or do not effectively capture the interrelations between drugs; these limitations hinder the methods from accurately predicting drug-pair interactions. In this paper, we present the effectiveness of variational graph autoencoders (VGAE) in modeling latent node representations on multimodal networks. Our approach can produce flexible latent spaces for each node type of the multimodal graph; the embeddings are used later for predicting links among node pairs under different edge types. To further enhance the models' performance, we suggest a new method that concatenates Morgan fingerprints, which capture the molecular structures of each drug, with their latent embeddings before preceding them to the decoding stage for link prediction. Our proposed model shows competitive results on three multimodal networks: (1) a multimodal graph consisting of drug and protein nodes, (2) a multimodal graph constructed from a subset of the DrugBank database involving drug nodes under different interaction types, and (3) a multimodal graph consisting of drug and cell line nodes. Our source code is publicly available at \url{https://github.com/HySonLab/drug-interactions}.
\end{abstract}


\section{Introduction}
In recent studies, estimates of the average Research $\&$ Development cost per new drug range from less than $\$1$ billion to more than $\$2$ billion per drug. Creating a new drug is not just incredibly expensive but also time-consuming. The full research, development and approval process can last from 12 to 15 years. Meanwhile the world has been facing new threats from epidemic diseases caused by new and constantly evolving viruses (e.g, COVID-19 caused by SARS-CoV-2). Therefore, it is essential to reuse/repurpose existing drugs and discover new ways of combining drugs (i.e. polypharmacy) in order to treat never-seen-before diseases. Most approaches to tackling drug repurposing problems estimate drug side effects and drug responses to cell lines to provide important information about the use of those drugs. In particular, polypharmacy is the practice of treating complex diseases or co-existing health conditions by using combinations of multiple medications \citep{bansal2014community}. Understanding drug-drug interactions (DDIs) is pivotal in predicting the potential side effects of multiple co-administered medications. It is, however, unfeasible to render clinical testing of all drug combinations due to the tremendous number of relations between drug pairs or drugs and their targets (e.g., proteins).
\vskip 1em
Machine learning models have been widely applied to provide tractable solutions in predicting potential drug interactions. Data-driven techniques (e.g., machine learning or deep learning) have proved their capabilities for dealing with complex patterns in data. Such methods have produced remarkable results in various fields, such as computer vision, natural language processing, and audio processing. Therefore, it is reasonable to employ machine learning approaches to investigate the interactions among complex combinations of medications. Previous work on DDIs \citep{gottlieb2012indi, vilar2012drug, cheng2014machine} used various hand-crafted drug features (e.g., chemical properties or Morgan fingerprints) to compute the similarity of each drug pair, and drug-drug interactions can be predicted based on the drug-pair proximity scores. Nevertheless, using fixed drug representations can result in sub-optimal outcomes since they do not sufficiently capture the complex interrelations of drugs. Recent approaches rely on graph neural networks (GNNs) \citep{battaglia2018relational} to directly learn drug and protein representations from structural data (i.e. graphs) augmented with additional information for node features. \cite{10.1093/bioinformatics/bty294} proposed a graph autoencoder to predict polypharmacy side effects on a multimodal graph consisting of drug and protein nodes with various edge types. Similarly, \cite{yin2022deepdrug} introduced DeepDrug combining graph convolutional neural networks (GCN) \citep{Kipf:2017tc} and convolutional neural networks (CNN)\citep{lecun1995convolutional} to learn the structural and sequential representations of drugs and proteins. In addition, \cite{wang2022deepdds} suggested using several graph attention (GAT) \citep{velickovic2018graph} layers to learn on hand-crafted features of drugs to predict drug-pair interactions. Nyamabo
et al. \cite{nyamabo2021ssi} use multiple GAT layers to compute the drug representations and aggregate them using co-attention to produce final predictions. These mentioned GNN-based models effectively learn the latent representations of drugs and proteins, which can be used in several downstream tasks. 
\vskip 1em
We hypothesize that a multimodal network can provide more comprehensive information for learning drug representations than the aforementioned methods. Rather than only considering pair-wise associations of drugs, neural networks operating on multimodal graphs can acknowledge the connections of multiple drugs under a specific interaction type, resulting in better modeling of drug interactions (e.g., polypharmacy side effects in \citep{Zitnik2017}). In other words, each interaction type is modeled as a graph of drug nodes and their pair-wise relations, and the aggregation of their vectors can be regarded as an informative representation of this interaction. Predicting drug-drug or drug-target interactions can be regarded as a link prediction task on graphs. We argue that latent spaces produced by the aforementioned approaches are disjoint; hence, they are not capable of generating new links on some graph benchmarks. Therefore, we suggest using deep graph generative models to make the latent spaces continuous; as a result, they are superior to traditional graph autoencoders (GAE) in predicting new links on biomedical multimodal graphs. Instead of yielding deterministic latent representations, variational graph autoencoders (VGAE) \citep{kipf2016variational} use a probabilistic approach to compute the latent variables. In this work, we aim to use VGAE to learn the representations of drugs and proteins on a multimodal graph; then, we predict several node-pair interactions (e.g., drug-pair polypharmacy side effects, drug-cell line response, etc.) using the learned representations. We demonstrate that using VGAE can attain superior performance in predicting drug-drug interactions on multimodal networks, outperforming GAE-based architectures and hand-crafted feature-based methods. In addition, we leverage drug molecular representations (i.e. Morgan fingerprints) concatenated with drug latent representations before the decoding stage to further enhance the model's performance in predicting drug-pair polypharmacy side effects. Experiments are conducted on three different biomedical datasets. Our proposed approach shows promising quantitative results, compared with other methods. Moreover, visualizations are presented to provide some findings in modeling drug interactions by multimodal graphs and deep generative models. 

\section{Related Work}
\subsection{Link Prediction with Graph Neural Networks}
Graph neural networks (GNNs) are deep learning models that learn to generate representations on graph-structured data. Most GNN-based approaches use message-passing paradigms \citep{4700287, pmlr-v70-gilmer17a} wherein each node aggregates vectorized information from its neighbors and updates the messages to produce new node representations. \cite{Kipf:2017tc} introduced a scalable method that approximates first-order spectral graph convolutions, which are equivalent to one-hop neighborhood message-passing neural networks. Besides, \cite{velickovic2018graph} proposed graph attention networks (GATs) that use soft attention mechanisms to weigh the messages propagated by the neighbors. \cite{NIPS2017_5dd9db5e} introduced GraphSAGE which efficiently uses node attribute information to generate representations on large graphs in inductive settings. Link prediction is one of the key problems in graph learning. In link-level tasks, the network consists of nodes and an incomplete set of edges between them, and partial information on the graph are used to infer the missing edges. The task has a wide range of variants and applications such as knowledge graph completion \citep{10.5555/2886521.2886624}, link prediction on citation networks \citep{bojchevski2018deep}, and predicting protein-protein interactions\citep{Zitnik2017}. 

\subsection{Deep Generative Model on Graphs}
Deep generative models aim to generate realistic samples, which should satisfy properties in nature. Different from traditional methods that rely on hand-crafted features, data-driven approaches can be categorized into variational autoencoders (VAEs) \citep{kingma2013auto}, generative adversarial networks (GANs) \citep{NIPS2014_5ca3e9b1}, and autoregressive models. In the field of deep generative models on graphs, several works \citep{pmlr-v70-ingraham17a, NEURIPS2020_41d80bfc} aim at adding stochasticity among latents to make the models more robust to complicated data distributions. \cite{graphvae} proposed a regularized graph VAE model wherein the decoder outputs probabilistic graphs. In addition, \cite{netgan} introduced stochastic neural networks trained by Wassterstain GAN object to learn the distribution of random walk distribution on graphs. For autoregressive methods, \cite{gran} proposed an attention-based GNN network to generate graph nodes and associated edges after many decision steps in a generation process.

\subsection{Drug-Drug Interaction Modeling}
Previous studies predicted drug-drug interactions by measuring the relevance between two graphical representations of drug chemicals (i.e. 2D graphs).  In particular, \cite{nyamabo2021ssi} introduce a deep neural network stacked by four GAT layers and a co-attention layer to compute the interaction relevance of two drug chemical substructures. \cite{mrgnn} propose to learn drug representations in a multi-resolution architecture. Moreover, concatenations of drug representations calculated by GNN-based models are insufficient to predict drug-pair interactions as they do not capture joint drug-drug information, which is regarded as interrelationships among medications. To overcome this problem, \cite{co_attention} use a co-attention mechanism that allows their message-passing-based neural networks to exchange substructure information between two drugs, resulting in better representations of individual drugs. Additionally,  \cite{gognn} train a model to capture graph-graph interactions (i.e. interactions between drugs at the chemical structure level) through dual attention mechanisms operating in the view of a graph consisting of multiple graphs. \cite{10.1093/bioinformatics/bty294} model a multimodal network consisting of medical node entities (i.e. drugs and proteins) with their relations to extract informative knowledge for predicting drug-drug interactions.
\section{Method}
\subsection{Problem Setup}
Given a multimodal graph, our task is to predict whether two nodes of the same or different types are connected under a specific edge type. This can be regarded as a link-level prediction on graphs. In our work, the graph $G$ consists of two node types (e.g., drug and protein), and each side effect is represented as an edge type. In addition to polypharmacy interactions, $G$ also involves edges representing drug-protein and protein-protein interactions. Formally, let denote $G = (V, E, X)$, where $V = V_d \cup V_p$ is a union of two node sets of different types (i.e. $V_d$ is the set of drug nodes and $V_p$ is the set of protein nodes), $E$ is a set of edges, and $X = X_d \oplus X_p$ is a concatenated matrix denoting the node features of different node types. Each edge in $E$ is a triplet $(v_i, e, v_j)$ in which node $v_i$ interacts with node $v_j$ under a specific edge type $e$. In case of a binary classification setting, we perform negative sampling to generate non-edge labels to make the model robust to negative examples. Thus, the objective is to learn a function $f: E \rightarrow T$, where $f$ predicts the value of a particular triplet $(v_i, e, v_j)$; $T$ can be either $\{0,1\}$ or $\mathbb{R}$.

\subsection{Model Architecture}
\label{sec:sub2}
Our proposed model has three main components: an encoder, a latent encoder, and a decoder. 
\begin{figure}[h]
     \centering
     \begin{subfigure}[b]{0.4\textwidth}
         \centering
         \includegraphics[width =\textwidth]{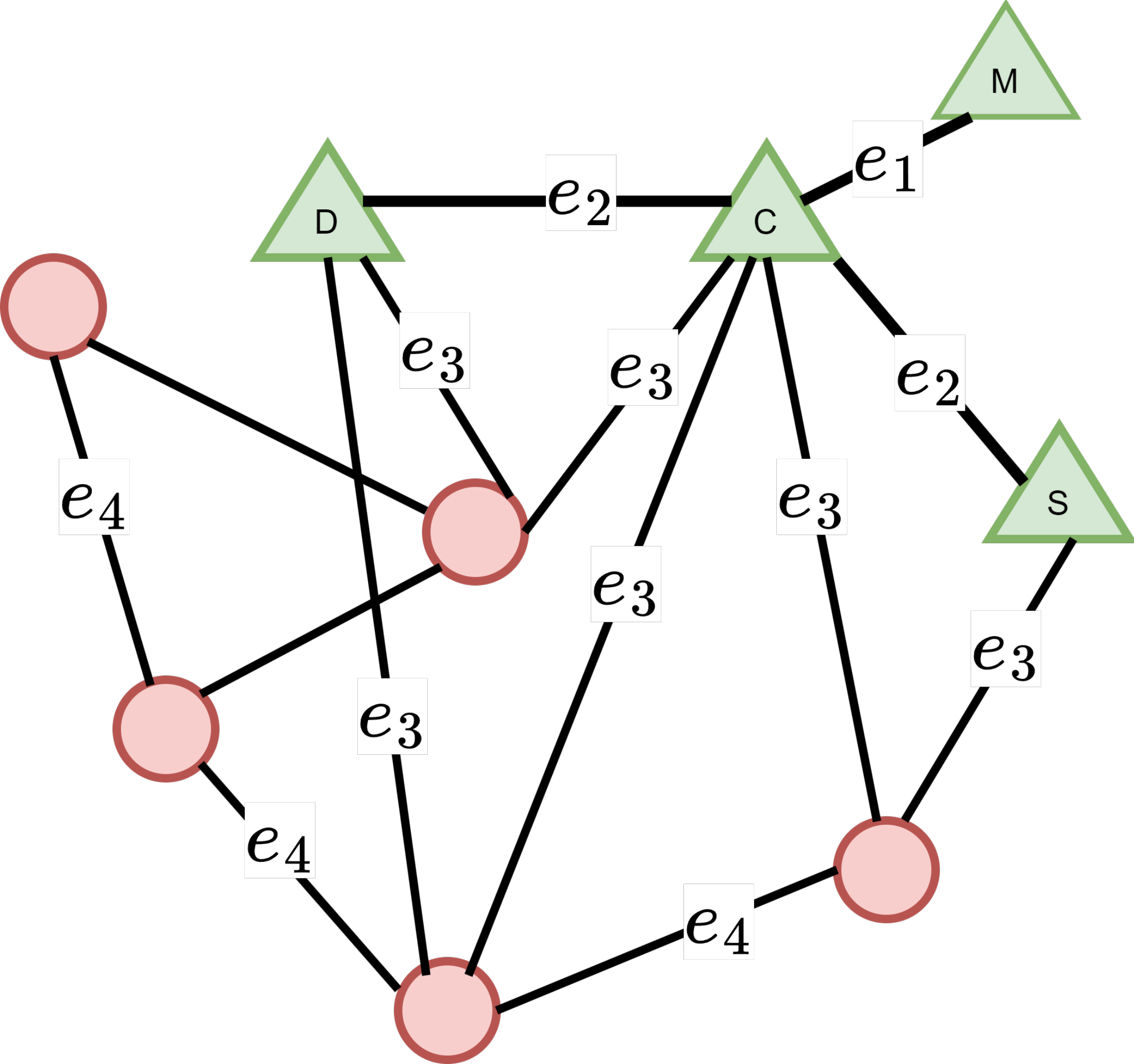}
         \caption{}
         \label{fig:graph}
     \end{subfigure}
     \hfill
     \begin{subfigure}[b]{0.5\textwidth}
         \centering
         \includegraphics[width =\textwidth]{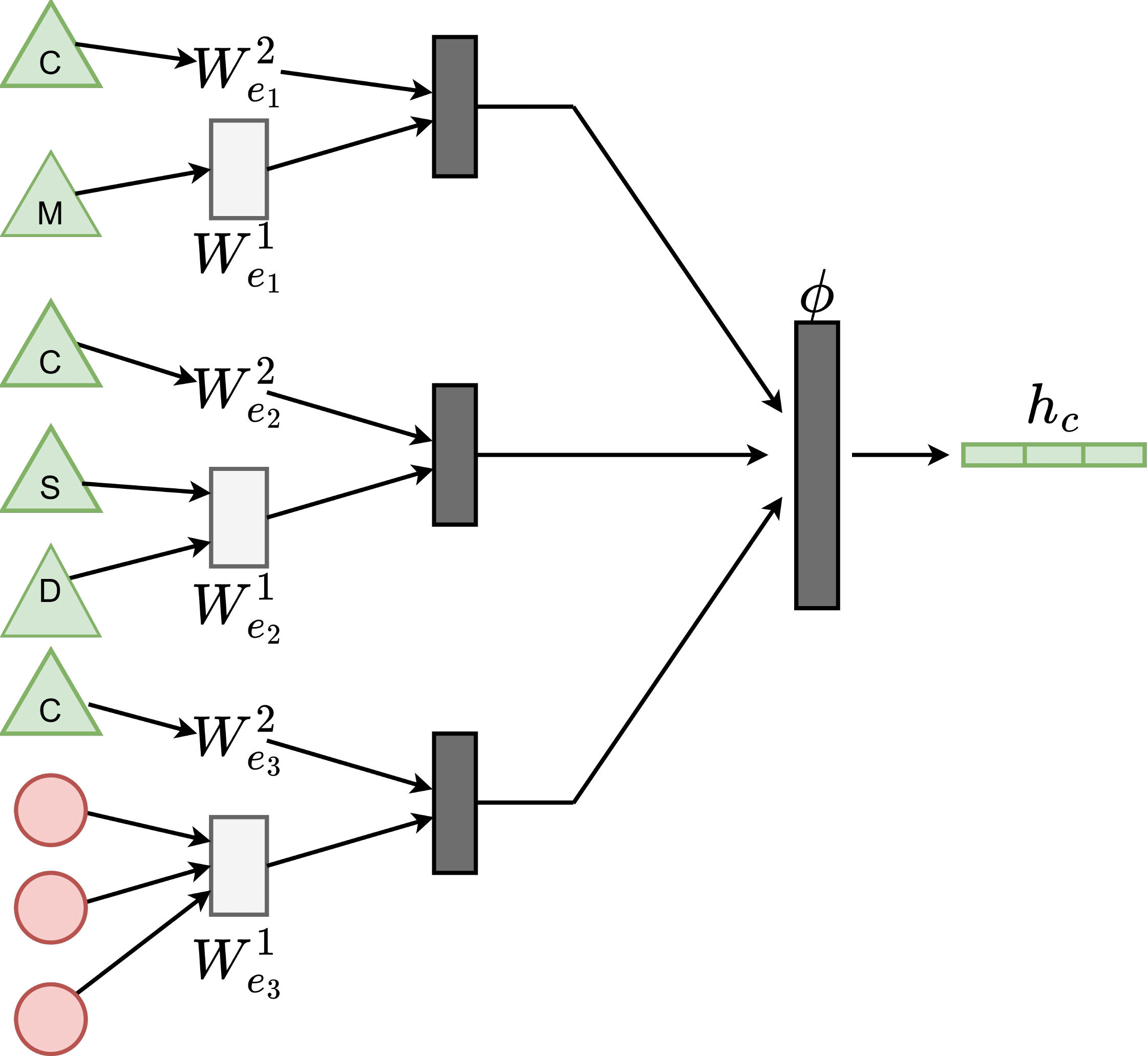}
         \caption{}
         \label{fig:encoder}
     \end{subfigure}
     \vskip 1.2em
        \caption{Overview of a biomedical multimodal graph and one graph convolution layer in our framework. (a) A graph consists of two node types (e.g., red nodes are protein and green are drug nodes) and five edge types $\{e_i\}$. (b) A layer of the encoder in which the representation of the source node $c$ is computed by aggregating its neighbors' information under different edge types. Black rectangles denote aggregation, whereas white rectangles indicate neural networks that share parameters across the nodes. $\{(W^1_{e_i}, W^2_{e_i})\}$ denote trainable weight matrices of different edge types $\{e_i\}$. $h_c$ can be moved to either a successive convolution layer or the latent encoder.}
        \label{fig:2}
\end{figure}

\subsubsection*{Encoder}
\label{latent_encoder} a two-layer graph convolutional neural network operating message passing scheme \cite{pmlr-v70-gilmer17a}  on $G$ and producing node embeddings for each node type (e.g., drugs and proteins). Each layer in the network has the following form: 
    \[h_i = \phi \big(\sum_{e}\sum_{j \in \mathcal{N}^i_e \cup \{i\}} W_{e}\frac{1}{\sqrt{c_i c_j}} x_j\big)\]
    where  $\mathcal{N}^i_e$ denotes the neighbor set of node $x_i$ under the edge type $e$.
    $W_{e} \in \mathbb{R} ^ {d_k \times d}$ is a edge-type specific transformation matrices that map $x_i \in \mathbb{R} ^ {d_i}$ and its neighbors $x_j \in \mathbb{R} ^ {d_j}$ into $d_k$-dimensional vector spaces, resulting in $h_i \in \mathbb{R}^{d_k}$. It is worth noting that $d_i$ and $d_j$ are not necessarily the same because $x_i$ and $x_j$ can be nodes of two different node types (i.e. $d_i = d_j$ when $x_i$ and $x_j$ are in the same node type). $c_i$ and $c_j$ indicates the degree node $i$ and $j$ on the network. Also, $\phi$ denotes a nonlinear activation function; we use rectified linear unit (ReLU) in our experiments. Figure \ref{fig:2} illustrates an overview of the encoder in our framework.

\thickmuskip=0mu 
 \subsubsection*{Latent Encoder}
\begin{figure}[h]
    \centering
    \includegraphics[scale=0.4]{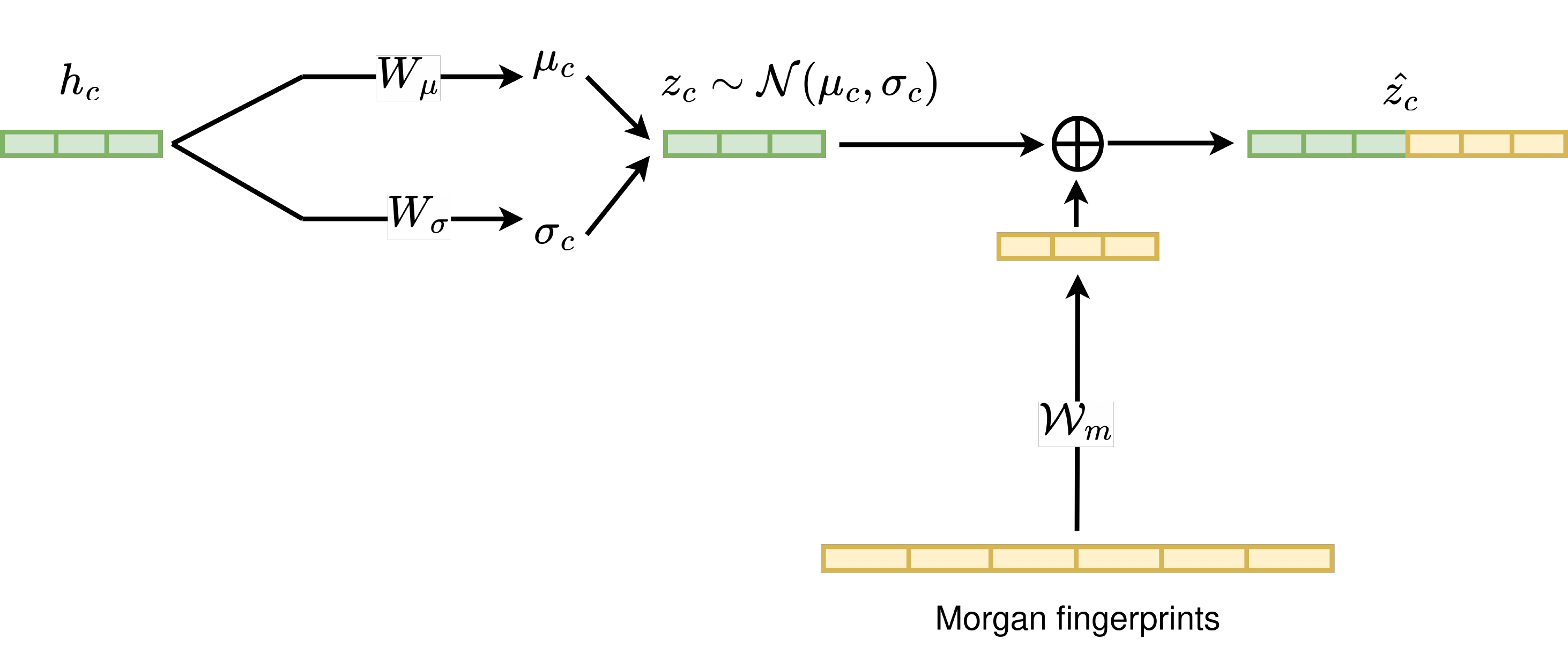}
    \caption{Overview of the latent encoder. Morgan fingerprints are concatenated with latent vector $z_c$ to improve the performance of polypharmacy side effects prediction. $\{W_\mu, W_\sigma, W_m$\} are trainable weight matrices. $\oplus$ denotes concatenation.} 
    \label{fig:3}
\end{figure}

 For each node type $v$, there are two multilayer perceptrons (MLPs) receiving the node embeddings from the encoder and computing the predicted mean $\mu$ and logarithm of the standard deviation $\log \sigma$ of the posterior latent distribution: 
    \[q_v (Z_v\mid X, E) = \prod_{i = 1}^{|V_v|}q_v(z^i_{v} \mid X, E)\]
    $q_v(z^i_v \mid X, E) = \mathcal{N}(z^i_v \mid \mu^i_v, \textrm{diag}((\sigma^i_v)^2)$ denotes the posterior distribution of a node of a specific node type. Here, $\mu_v$ and $\log \sigma _v$ are computed as follows:
    \[\mu_v = W^2_{\mu_v} \tanh(W^1_{\mu_v} h_v)\]
    \[\log \sigma_v = W^2_{\sigma_v} \tanh(W^1_{\sigma_v} h_v)\]
    where $W^i_{\mu_v} \in \mathbb{R} ^ {d_k \times d}$, $W^i_{\sigma_v} \in \mathbb{R} ^ {d_k \times d}$ are the weight matrices, $\mu_v$ and $\log \sigma_v$ are the matrices of mean vector $\mu^i_v$ and logarithm of standard deviation vector $\log \sigma^i_v$, respectively. Figure \ref{fig:3} demonstrates how we integrate molecular structures of drug nodes with their latent representation to improve the performance in side effect link prediction. 
 \subsubsection*{Decoder}
 In our experiments, we use two types of decoders including tensor factorization and Multilayer perception (MLP) for different purposes. A tensor-based decoder is used in the polypharmacy side effects experiments not only to make predictions but the on-diagonal entries of the tensor can also be used as representations for the drug-pair side effects.  On the other hand, a multilayer perceptron (MLP) are used in the two remaining experiments to provide better performances.
\paragraph{Tensor Factorization:}
 a tensor factorization using latent embeddings to predict the node interactions on $G$. We follow the approach proposed in \cite{10.1093/bioinformatics/bty294} to design the decoder: 
    \begin{equation*}
        g(v_i, e, v_j) = 
        \begin{cases}
            z^T_i D_e R D_e z_j & \text{if $v_i$ and $v_j$ are drugs} \\
            z^T_i M_e z_j & \text{if $v_i$ and $v_j$ are a protein and a drug, or vice versa.}
        \end{cases}
    \end{equation*}
    where $D_e, R, M_e \in \mathbb{R} ^ {d \times d}$ are learnable parameters. $R$ denotes the global matrix representing all drug-drug interactions among all polypharmacy side effects; $M_e$ is an edge-type-specific matrix modeling drug-protein and protein-protein relations.
    Also, $D_e$ is a diagonal matrix, and its on-diagonal entries model the significance factors of $z_i$ and $z_j$ in multiple dimensions under the side effect type $e$.
\paragraph{Multilayer perceptron:} a multilayer perceptron (MLP) can also be used to predict the links between nodes as follows:
\begin{equation*}
    g(v_i, e, v_j) = \texttt{MLP}(\texttt{concat}(z_i, z_j))
\end{equation*}
where $z_i$ and $z_j$ are two drug vectors sampled from the latent posterior distribution.
\vskip 0.5em
\noindent
Finally, the probability of edge $(v_i, e, v_j)$ is calculated via a sigmoid function $\sigma$.
    \[p_e(v_i, v_j) = \sigma(g(v_i, e, v_j))\]

\section{Experiments}
\label{sec:3}
\subsection{Polypharmacy side effects}

We conduct experiments on the dataset introduced in \cite{10.1093/bioinformatics/bty294}. It is a multimodal graph consisting of 19,085 protein and 645 drug nodes. There are three types of interaction drug-drug, drug-protein, and protein-protein. In particular, the dataset contains 964 commonly occurring polypharmacy side effects, resulting in 964 edge types of drug-drug interaction. Protein-protein and drug-protein interactions are regarded as two other edge types; therefore, the multimodal graph has 966 edge types in total.
\vskip 1em
We randomly split the edges into training, validation, and testing sets with a ratio of $8 : 1 : 1$. Then, the edges in the training edge set are randomly divided into $80\%$ edges for message passing and $20\%$ edges for supervision. We use Adam optimizer to minimize the loss function $L$ shown in Equation \ref{eq:1} in 300 epochs with a learning rate of $0.001$. We run the experiments with six different random seeds. 

\begin{equation}
    L = \sum_{(v_i, e, v_j) \in E}-\log p_e(v_i, v_j) - \mathbb{E}_{v_n \sim P_e(v_i)}[ 1 - \log p_e(v_i, v_n)] - \sum_{v} \lambda_v \mathbb{D}_{KL}(q_v(Z_v\mid X, E)\mid p_v(Z_v))
    \label{eq:1}
\end{equation}
where $p_v(Z_v)$ denotes the latent space's prior distribution of the node type $v$. The first term of $L$ denotes the cross-entropy loss of the probabilities of positive and negative edges which is sampled by choosing a random node $v_n$ for each node $v_i$ under a specific edge type $e$, while the second term is the weighted sum of KL-divergence of each node type $v$; $\{\lambda_v\}$ are hyper-parameters. In our experiments, $\lambda_d$ and $\lambda_p$ equal to $0.9$ and $0.9$ for drug and protein nodes, respectively. Finally, we use one-hot representations indicating node indices as features for both drug and protein nodes.  

\subsection{DrugBank Drug-Drug Interaction}
We run experiments on the DrugBank dataset used in \citep{drug_food}. In this dataset, 191,400 drug-pair interactions among 1,704 drugs are categorized into 86 types and split into training, validation, and testing subsets with a ratio of 6:2:2 respectively. We use the SMILES string representations of drug nodes to compute their 2,048-dimension Morgan fingerprints feature vectors which indicate the molecular structures of the medications. 
\vskip 1em
In this task, we build a multimodal network in which drugs are nodes and their interactions are edge types, which is similar to the polypharmacy prediction task. The nodes are attributed to their Morgan fingerprint representations. To the best of our knowledge, our work is the first attempt to approach the DrugBank dataset in this way. Morgan fingerprint vectors are good molecular features of individual drugs, and aggregating drug information under different edge types can produce superior drug interrelationships. In our experiments, we construct the multimodal network using drug pairs from the training set. Our VGAE model learns to predict the interactions of a drug pair under different edge types by minimizing the loss function in equation \ref{eq:1}. In this task, we use multiplayer perceptron as the decoder since the main objective is to predict different interactions between two medications rather than modeling their co-occurring effects.

\subsection{Anticancer drug response}
In addition to drug-pair polypharmacy side effects, we evaluate our approach for the anticancer drug response prediction problem; the task is  to predict the interactions between the drug and cell line on a multimodal network. 
Integrated information between drugs and cell lines are an effective approach to calculating anticancer drug responses using computational methods; various type of information can be exploited such as drug chemical structures, gene expression profiles, etc., to provide more accurate predictions. In this work, we evaluate the performance of VGAE on the CCLE dataset proposed in \citep{ahmadi2020adrml} which contains comprehensive information of drugs and cell lines. Three types of pairwise similarities are provided for each modality, resulting in nine types of combination used in computation among drug and cell line pairs; in addition, pairwise logarithms of the half-maximal inhibitory concentration (IC50) scores between drugs and cell lines are given as values to be predicted. The combinations are detailed in Table \ref{tab:3}.

To apply VGAE in predicting the IC50 scores among drug-cell line pairs, we construct an undirected multimodal graph consisting of drug and cell line nodes with the similarity scores indicating the edge weights under three types of relation, including drug-drug, cell line-cell line, and drug-cell line. The problem is regarded as a link prediction task in which a number of edges connecting drug and cell line nodes are masked, and the objective is to compute real values $w_e$ denoting the weights of these missing links. The implementation of VGAE is almost the same as detailed in \ref{sec:sub2} with modifications of the decoding stage in which representations of two nodes are concatenated before being preceded to a simple multilayer perceptron with two hidden layers of size 16, ReLU nonlinearity, and a linear layer on top to predict a final score denoting the edge weights among them. 

We randomly split the edges with a ratio of 7:1:2 for training, validation, and testing purposes, respectively. In this task, the models are trained to reconstruct a weighted adjacency matrix and regularized by KL-divergence of the node latent distributions with hyperparameters $\lambda_d$ and $\lambda_c$ denoting the coefficients for drug and cell nodes, respectively; $\lambda_d = \lambda_c = 0.001$ in our experiments. We train the models in 500 epochs with a learning rate of 0.01 to minimize the loss function $L$ as follows: 
\begin{equation}
    L = \sum_{(v_i, e, v_j) \in E} (\widehat{s_e}(v_i, v_j) - s_e(v_i, v_j))^2
    - \sum_{v} \lambda_v \mathbb{D}_{KL}(q_v(Z_v\mid X, E)\mid p_v(Z_v))
    \label{eq:2}
\end{equation}
where $\widehat{s_e}(v_i, v_j)$ indicates the predicted edge weights between node $v_i$ and $v_j$, whereas $s_e(v_i, v_j)$ are their ground truths. Finally, experiments are run with 20 different random seeds.

\section{Comparative Results}
\subsection{Polypharmacy side effects prediction}
We compare the performance of VGAE to alternative approaches. In addition to VGAE, we also implement a graph autoencoder and train the model in the same setting detailed in Sec. \ref{sec:3}. The baseline results are taken from \citep{10.1093/bioinformatics/bty294} including: RESCAL Tensor Factorization \citep{nickel2011three}, DEDICOM Tensor Factorization \citep{papalexakis2016tensors}, DeepWalk Neural Embeddings \citep{perozzi2014deepwalk, zong2017deep}, Concatenated Drug Features, and Decagon \citep{10.1093/bioinformatics/bty294}. It is worth noting that Decagon is also a GAE-based model with two GCN layers, yet \cite{10.1093/bioinformatics/bty294} use side information (e.g., side effects of individual drugs) as additional features for drug nodes. By contrast, our VGAE and GAE are trained on one-hot representations for drug and protein nodes. To examine the effects of Morgan fingerprints in drug-drug interaction decoding, we augment the Morgan fingerprint information into the latent vectors of drug nodes before preceding them to tensor-based decoders to produce final predictions.
\vskip 1em
Table \ref{tab:2} shows the results of all approaches in predicting the polypharmacy side effects. The models are evaluated based on three metrics, including the area under the ROC curve (AUROC), the area under the precision-recall curve (AUPRC), and average precision at 50 (AP@50). The scores reveal that VGAE without augmenting Morgan fingerprints outperforms traditional tensor-based approaches by a large margin, resulting in up to $24.8 \%$ (AUROC), $32.0 \%$ (AUPRC), and $40.3 \%$ (AP@50). We also compare VGAE with two machine-learning-based methods. The model achieves a $25.8 \%$ gain over DeepWalk neural embeddings and $18.0 \%$ over Concatenated drug features in AP@50 scores. Furthermore, albeit trained on one-hot feature vectors, vanilla VGAE can achieve competitive performance with Decagon, outperforming the baseline by $6.0\%$ (AP@50). This indicates the effectiveness of VGAE in recommending potential side effects in a featureless drug-protein multimodal network. Finally, VGAE with additional molecular information (i.e. VGAE + MFs) at the decoding stage provides the best performance across the approaches. The method achieves the highest scores among all three metrics and outperforms the others by a large margin. The results reveal that using molecular fingerprints is a simple yet effective solution in predicting drug-drug interactions as suggested in \citep{long500molecular}.
\begin{table}[ht]
	\centering
	\caption{Average performance on polypharmacy link prediction across all side effects}
	\vskip 1.2em
	\label{tab:sample}
	\begin{tabular}{cccc}
		\toprule
		Method & AUROC & AUPRC & AP@50 \\
		\midrule
		RESCAL  & 0.693 & 0.613 & 0.476 \\
		DEDICOM  & 0.705 & 0.637 & 0.567 \\
		DeepWalk & 0.761 & 0.737 & 0.658 \\
		Concatenation & 0.793 & 0.764 & 0.712 \\
	    Decagon & 0.872 & 0.832 & 0.803 \\
	    GAE & 0.893 $\pm$ 0.002 & 0.862 $\pm$ 0.003 & 0.819 $\pm$ 0.006 \\
	    \midrule
		VGAE (ours) & 0.905 $\pm$ 0.001 & 0.880 $\pm$ 0.001 & 0.853 $\pm$ 0.005 \\
		VGAE + MFs (ours) & \textbf{0.944 $\pm$ 0.005} & \textbf{0.926 $\pm$ 0.005} & \textbf{0.920 $\pm$ 0.004} \\
		\bottomrule
	\end{tabular}
	\label{tab:2}
\end{table}
\paragraph{Visualization of side effect embeddings}
\begin{figure}
    \centering
    \captionsetup{justification=centering}
    \includegraphics[width = 0.65\textwidth]{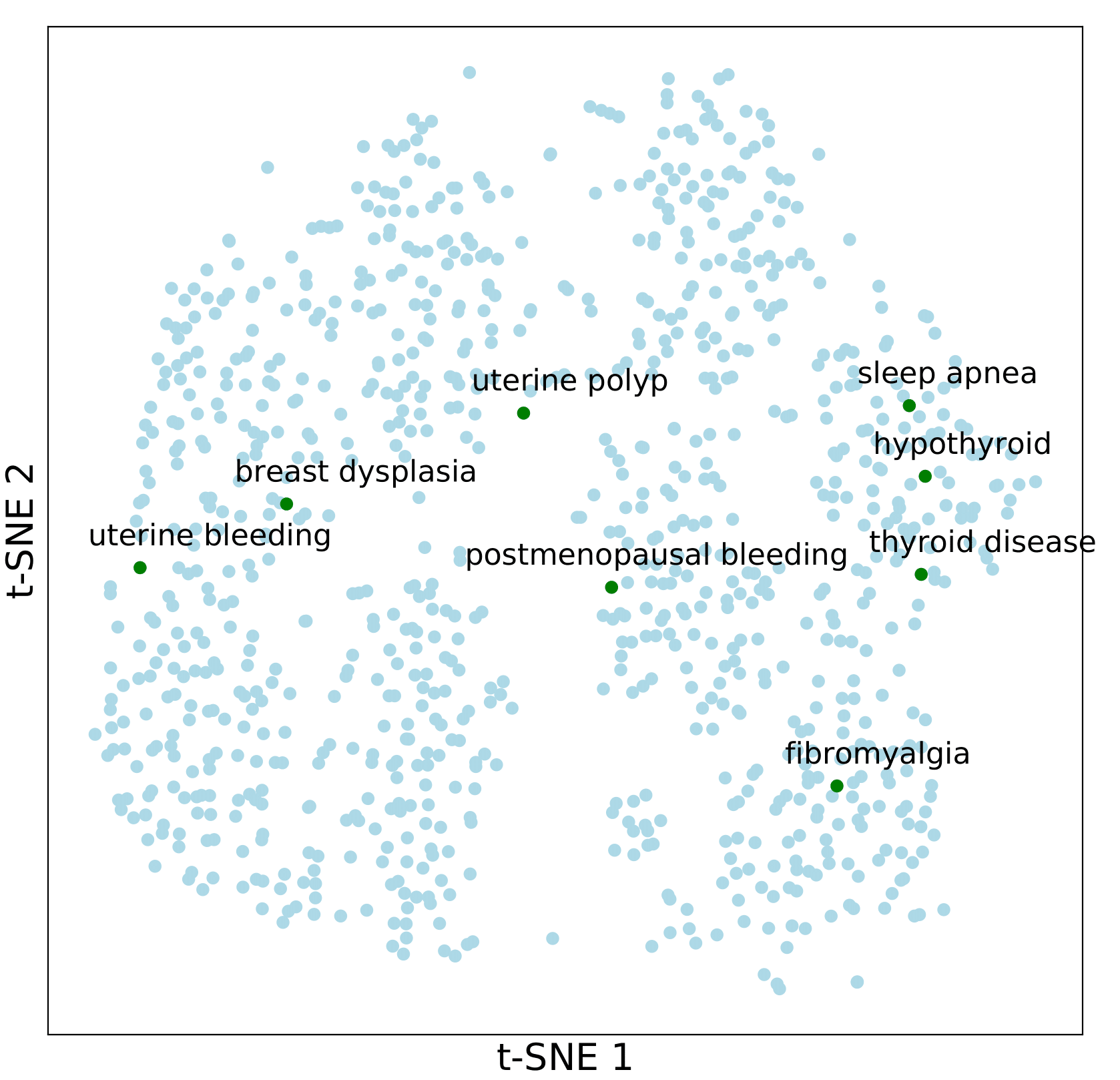}
    \caption{Polypharmcy side effects embeddings}
    \label{fig:1}
\end{figure}

To further demonstrate the capability of VGAE in learning to model drug-pair polypharmacy side effects, we plot their representations in 2D dimension using the t-SNE \cite{van2008visualizing} method. The embedding of each side effect is derived by taking on-diagonal entries of the tensor $D_e$ to create a $d$-dimensional vector which is accordingly projected into 2D space. Figure \ref{fig:1} illustrates a vector space in which  side effects representations establish a clustering structure. 

\begin{figure}
    \centering
    \includegraphics[scale=0.6]{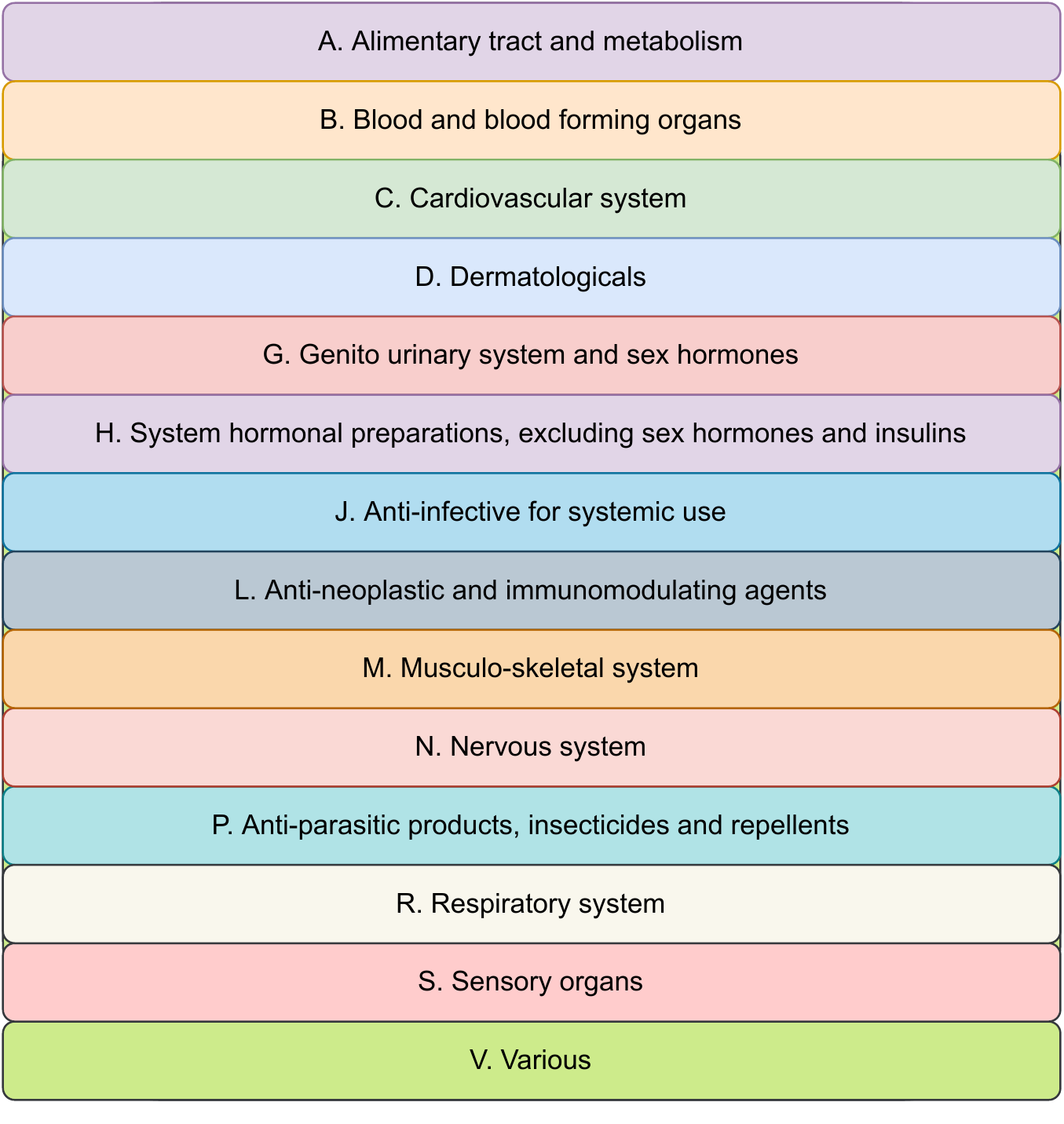}
    \caption{Fourteen groups at Level 1 in Anatomical Therapeutic Chemical classification system}
    \label{fig:atc}
\end{figure}
\paragraph{Visualization of drug embeddings}
\begin{figure}[ht]
    \centering
    \includegraphics[scale=0.4]{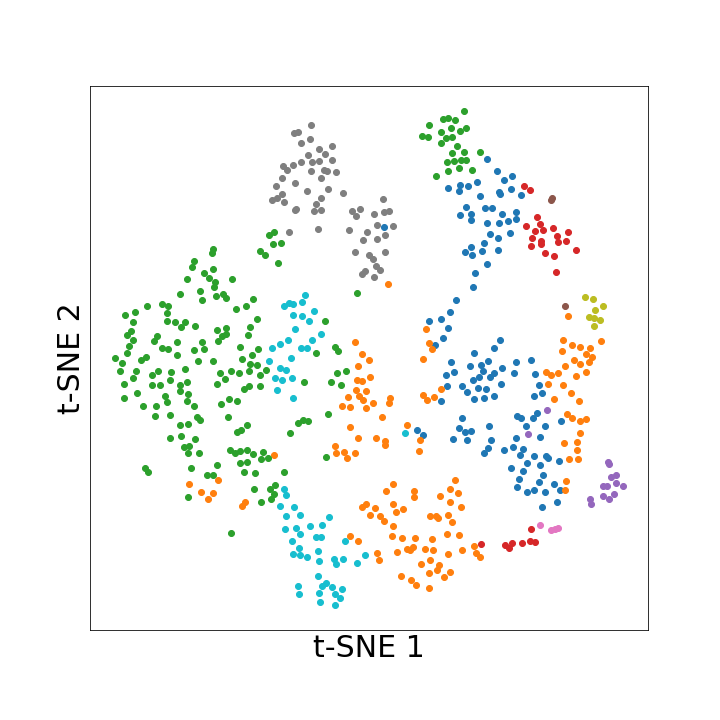}
    \caption{Visualization of drug embeddings indicates how drugs are clustered into groups}
    \label{fig:latent}
\end{figure}
We run a spectral algorithm to find a clustering structure of the latent drug space produced by the proposed model. Specifically, the number of clusters is determined based on the Anatomical Therapeutic Chemical (ATC) classification system \cite{atc} in which medications are divided into groups at five different levels. At level 1, drugs are classified into 14 pharmacological groups. Figure \ref{fig:atc} demonstrates the drug classification at level 1 of ATC. In our framework, therapeutic knowledge of drugs is extracted from a biomedical multimodal graph and Morgan fingerprints can represent their chemical structures. Therefore, we hypothesize that drug latent vectors combined with both pharmacological and molecular structures can derive informative drug clustering results that are useful for classification systems (e.g., ATC) to determine the functions of new medications.
The clustering latent space shown in Figure \ref{fig:latent} demonstrates a promising result, suggesting new directions for further work in learning to cluster drugs or diseases by exploring their joint representations on multimodal networks.

\subsection{DrugBank Drug-Drug Interaction}
\begin{table}[ht]
	\centering
	\caption{\centering Comparisons on the DrugBank dataset}
	\vskip 1.2em
	\begin{tabular}{cccc}
		\toprule
		Method & ACC & AUROC & AUPRC \\
		\midrule
		  DeepDDI  & 0.9315 & \textbf{0.9976} & 0.9319 \\
		MR-GNN  & 0.9406 & \textbf{0.9976} & 0.9410 \\
		GoGNN & 0.8478 & 0.9163 & 0.9001 \\
		  MHCADDI & 0.7850 & 0.8633 & 0.8938 \\
	    SSI-DDI & 0.9447 & 0.9838 & 0.9814 \\
	    \midrule
		VGAE (ours) & \textbf{0.9788} & 0.9938 & \textbf{0.9914} \\
		\bottomrule
	\end{tabular}
	\label{tab:6}
\end{table}
We report the experimental results on three metrics: accuracy (ACC), the area under the ROC curve (AUROC), and the area under the precision-recall curve (AUPRC). The baselines are taken from \cite{nyamabo2021ssi}, including DeepDDI \cite{drug_food}, MR-GNN \cite{mrgnn}, GoGNN \cite{gognn}, MHCADDI \cite{co_attention}, and SSI-DDI \cite{nyamabo2021ssi}. Table \ref{tab:6} summarizes the results of different methods evaluated on the DrugBank dataset. According to the table, our proposed method surpasses the competitors in ACC and AUPRC and achieves comparable scores in AUROC. VGAE attains the highest accuracy (ACC) at $97.88\%$, outperforming SSI-DDI by $3.5\%$. Similarly, the AUPRC produced by VGAE is relatively higher than SSI-DDI by $1.0\%$.

\subsection{Anticancer drug response prediction}
In this task, the models are evaluated using four criteria including root mean square error (RMSE), coefficient of determination ($R^2$), Pearson correlation coefficient (PCC), and fitness which is computed as: 
\[fitness \,=\, R^2 + PCC - RMSE\]

Table \ref{tab:3} shows the experimental results of drug response prediction on nine combinations of pairwise similarities of drugs and cell lines. Drug pairwise similarities are calculated via three different properties, including chemical structures (Chem), target protein (Target), and KEGG pathways (KEGG). For cell lines, their similarities are determined by mutation (Mutation), gene expressions (GE), and copy number variation (CNV).
In particular, VGAE trained on copy number variation with KEGG-pathways as similarity scores for cell lines and drugs achieve the best performance, yielding the lowest RMSE at 0.46 $\pm$ 0.02 and highest scores in $R^2$ = 0.67 $\pm$ 0.03, PCC = 0.85 $\pm$ 0.01, and fitness = 1.05 $\pm$ 0.06. Moreover, Table \ref{tab:4} demonstrates the comparisons between VGAE and other baselines which are taken from \citep{ahmadi2020adrml} including CDRscan
\citep{chang2018cancer}, CDCN \citep{wei2019comprehensive}, SRMF \citep{suphavilai2018predicting}, CaDRRes \citep{wang2017improved}, ADRML \citep{ahmadi2020adrml}, and
k-nearest neighbors (KNN). The results reveal that compared with the baselines, VGAE has competitive performance, especially in the fitness score determining the effectiveness of the approaches in the drug response prediction task. 
\begin{table}
	\centering
	\caption{Performance of VGAE on various types of similarities of drugs and cell lines}
	\vskip 1em
	\begin{tabular}{cccccc}
		\toprule
		Cell  & Drug & RMSE $\downarrow$ & $R^2$ $\uparrow$ & PCC $\uparrow$ & fitness $\uparrow$ \\
		\midrule
		
		Mutation & Chem & 0.48 $\pm$ 0.02 & 0.64 $\pm$ 0.04 & 0.84 $\pm$ 0.01 & 1.00 $\pm$ 0.06 \\
		GE & Chem & 0.47 $\pm$ 0.01 & 0.66 $\pm$ 0.03 & 0.85 $\pm$ 0.00 & 1.03 $\pm$ 0.05 \\
		CNV  & Chem & 0.47 $\pm$ 0.02 & 0.65 $\pm$ 0.03 & 0.84 $\pm$ 0.01 & 1.02 $\pm$ 0.05 \\
		Mutation & Target & 0.48 $\pm$ 0.02 & 0.66 $\pm$ 0.03 & 0.84 $\pm$ 0.01 & 1.01 $\pm$ 0.05 \\
		GE & Target & 0.47 $\pm$ 0.02 & 0.66 $\pm$ 0.03 & 0.84 $\pm$ 0.01 & 1.03 $\pm$ 0.05 \\
		CNV & Target & 0.48 $\pm$ 0.01 & 0.65 $\pm$ 0.03 & 0.84 $\pm$ 0.01 & 1.02 $\pm$ 0.05 \\
		Mutation & KEGG & 0.47 $\pm$ 0.02 & 0.66 $\pm$ 0.03 & 0.85 $\pm$ 0.01 & 1.03 $\pm$ 0.06 \\
		GE & KEGG & 0.47 $\pm$ 0.02 & 0.65 $\pm$ 0.04 & 0.84 $\pm$ 0.01 & 1.03 $\pm$ 0.08 \\
		CNV & KEGG & \textbf{0.46 $\pm$ 0.02} & \textbf{0.67 $\pm$ 0.03} & \textbf{0.85 $\pm$ 0.01} & \textbf{1.05 $\pm$ 0.06} \\
		\bottomrule
	\end{tabular}
	\label{tab:3}
\end{table}

\begin{table}
	\centering
	\caption{Methods' performance in predicting anticancer drug response}
	\vskip 1em
	\begin{tabular}{ccccc}
		\toprule
		Method  & RMSE $\downarrow$ & $R^2$ $\uparrow$ & PCC $\uparrow$ & fitness $\uparrow$ \\
		\midrule
		ADRML  & 0.49 & \textbf{0.68} & \textbf{0.85} & 1.04 \\
		CDRscan  & 0.76 & 0.67 & 0.83 & 0.74\\
		CDCN  & 0.48 & 0.67 & 0.83 & 1.02 \\
		SRMF  & \textbf{0.25} & 0.40 & 0.80 & 0.95 \\
	    CaDRRes & 0.53 & 0.31 & 0.52 & 0.3 \\
	    KNN  &  0.56 & 0.57 & 0.78 & 0.79 \\
	    \midrule 
	    VGAE (ours) & 0.46 $\pm$ 0.02 & 0.67 $\pm$ 0.03 & \textbf{0.85 $\pm$ 0.01} & \textbf{1.05 $\pm$ 0.06} \\
		\bottomrule
	\end{tabular}
	\label{tab:4}
\end{table}
\newpage
\section{Conclusion}
In this work, we evaluate the effectiveness of variational graph autoencoders in predicting potential polypharmacy side effects on multimodal networks. The results reveal that VGAE trained on one-hot feature vectors outperforms other approaches. Moreover, augmenting Morgan fingerprints before the decoding stage helps boost the performance of VGAE. This suggests further examination of the use of molecular fingerprints in drug-drug interaction problems.



\bibliography{sample}

\begin{thebibliography}{42}
\providecommand{\natexlab}[1]{#1}
\providecommand{\url}[1]{\texttt{#1}}
\expandafter\ifx\csname urlstyle\endcsname\relax
  \providecommand{\doi}[1]{doi: #1}\else
  \providecommand{\doi}{doi: \begingroup \urlstyle{rm}\Url}\fi

\bibitem[atc()]{atc}
Atc/ddd index 2022.
\newblock URL \url{https://www.whocc.no/atc_ddd_index/}.

\bibitem[Ahmadi~Moughari and Eslahchi(2020)]{ahmadi2020adrml}
Fatemeh Ahmadi~Moughari and Changiz Eslahchi.
\newblock Adrml: anticancer drug response prediction using manifold learning.
\newblock \emph{Scientific reports}, 10\penalty0 (1):\penalty0 1--18, 2020.

\bibitem[Bansal et~al.(2014)Bansal, Yang, Karan, Menden, Costello, Tang, Xiao,
  Li, Allen, Zhong, et~al.]{bansal2014community}
Mukesh Bansal, Jichen Yang, Charles Karan, Michael~P Menden, James~C Costello,
  Hao Tang, Guanghua Xiao, Yajuan Li, Jeffrey Allen, Rui Zhong, et~al.
\newblock A community computational challenge to predict the activity of pairs
  of compounds.
\newblock \emph{Nature biotechnology}, 32\penalty0 (12):\penalty0 1213--1222,
  2014.

\bibitem[Battaglia et~al.(2018)Battaglia, Hamrick, Bapst, Sanchez-Gonzalez,
  Zambaldi, Malinowski, Tacchetti, Raposo, Santoro, Faulkner,
  et~al.]{battaglia2018relational}
Peter~W Battaglia, Jessica~B Hamrick, Victor Bapst, Alvaro Sanchez-Gonzalez,
  Vinicius Zambaldi, Mateusz Malinowski, Andrea Tacchetti, David Raposo, Adam
  Santoro, Ryan Faulkner, et~al.
\newblock Relational inductive biases, deep learning, and graph networks.
\newblock \emph{arXiv preprint arXiv:1806.01261}, 2018.

\bibitem[Bojchevski and Günnemann(2018)]{bojchevski2018deep}
Aleksandar Bojchevski and Stephan Günnemann.
\newblock Deep gaussian embedding of graphs: Unsupervised inductive learning
  via ranking.
\newblock In \emph{International Conference on Learning Representations}, 2018.
\newblock URL \url{https://openreview.net/forum?id=r1ZdKJ-0W}.

\bibitem[Bojchevski et~al.(2018)Bojchevski, Shchur, Z{\"u}gner, and
  G{\"u}nnemann]{netgan}
Aleksandar Bojchevski, Oleksandr Shchur, Daniel Z{\"u}gner, and Stephan
  G{\"u}nnemann.
\newblock {N}et{GAN}: Generating graphs via random walks.
\newblock In Jennifer Dy and Andreas Krause, editors, \emph{Proceedings of the
  35th International Conference on Machine Learning}, volume~80 of
  \emph{Proceedings of Machine Learning Research}, pages 610--619. PMLR, 10--15
  Jul 2018.
\newblock URL \url{https://proceedings.mlr.press/v80/bojchevski18a.html}.

\bibitem[Chang et~al.(2018)Chang, Park, Yang, Lee, Lee, Kim, Jung, and
  Shin]{chang2018cancer}
Yoosup Chang, Hyejin Park, Hyun-Jin Yang, Seungju Lee, Kwee-Yum Lee, Tae~Soon
  Kim, Jongsun Jung, and Jae-Min Shin.
\newblock Cancer drug response profile scan (cdrscan): a deep learning model
  that predicts drug effectiveness from cancer genomic signature.
\newblock \emph{Scientific reports}, 8\penalty0 (1):\penalty0 1--11, 2018.

\bibitem[Cheng and Zhao(2014)]{cheng2014machine}
Feixiong Cheng and Zhongming Zhao.
\newblock Machine learning-based prediction of drug--drug interactions by
  integrating drug phenotypic, therapeutic, chemical, and genomic properties.
\newblock \emph{Journal of the American Medical Informatics Association},
  21\penalty0 (e2):\penalty0 e278--e286, 2014.

\bibitem[Deac et~al.(2019)Deac, Huang, Velickovic, Li{\`{o}}, and
  Tang]{co_attention}
Andreea Deac, Yu{-}Hsiang Huang, Petar Velickovic, Pietro Li{\`{o}}, and Jian
  Tang.
\newblock Drug-drug adverse effect prediction with graph co-attention.
\newblock \emph{CoRR}, abs/1905.00534, 2019.
\newblock URL \url{http://arxiv.org/abs/1905.00534}.

\bibitem[Gilmer et~al.(2017)Gilmer, Schoenholz, Riley, Vinyals, and
  Dahl]{pmlr-v70-gilmer17a}
Justin Gilmer, Samuel~S. Schoenholz, Patrick~F. Riley, Oriol Vinyals, and
  George~E. Dahl.
\newblock Neural message passing for quantum chemistry.
\newblock In Doina Precup and Yee~Whye Teh, editors, \emph{Proceedings of the
  34th International Conference on Machine Learning}, volume~70 of
  \emph{Proceedings of Machine Learning Research}, pages 1263--1272. PMLR,
  06--11 Aug 2017.
\newblock URL \url{https://proceedings.mlr.press/v70/gilmer17a.html}.

\bibitem[Goodfellow et~al.(2014)Goodfellow, Pouget-Abadie, Mirza, Xu,
  Warde-Farley, Ozair, Courville, and Bengio]{NIPS2014_5ca3e9b1}
Ian Goodfellow, Jean Pouget-Abadie, Mehdi Mirza, Bing Xu, David Warde-Farley,
  Sherjil Ozair, Aaron Courville, and Yoshua Bengio.
\newblock Generative adversarial nets.
\newblock In Z.~Ghahramani, M.~Welling, C.~Cortes, N.~Lawrence, and K.Q.
  Weinberger, editors, \emph{Advances in Neural Information Processing
  Systems}, volume~27. Curran Associates, Inc., 2014.
\newblock URL
  \url{https://proceedings.neurips.cc/paper/2014/file/5ca3e9b122f61f8f06494c97b1afccf3-Paper.pdf}.

\bibitem[Gottlieb et~al.(2012)Gottlieb, Stein, Oron, Ruppin, and
  Sharan]{gottlieb2012indi}
Assaf Gottlieb, Gideon~Y Stein, Yoram Oron, Eytan Ruppin, and Roded Sharan.
\newblock Indi: a computational framework for inferring drug interactions and
  their associated recommendations.
\newblock \emph{Molecular systems biology}, 8\penalty0 (1):\penalty0 592, 2012.

\bibitem[Hamilton et~al.(2017)Hamilton, Ying, and Leskovec]{NIPS2017_5dd9db5e}
Will Hamilton, Zhitao Ying, and Jure Leskovec.
\newblock Inductive representation learning on large graphs.
\newblock In I.~Guyon, U.~Von Luxburg, S.~Bengio, H.~Wallach, R.~Fergus,
  S.~Vishwanathan, and R.~Garnett, editors, \emph{Advances in Neural
  Information Processing Systems}, volume~30. Curran Associates, Inc., 2017.
\newblock URL
  \url{https://proceedings.neurips.cc/paper/2017/file/5dd9db5e033da9c6fb5ba83c7a7ebea9-Paper.pdf}.

\bibitem[Ingraham and Marks(2017)]{pmlr-v70-ingraham17a}
John Ingraham and Debora Marks.
\newblock Variational inference for sparse and undirected models.
\newblock In Doina Precup and Yee~Whye Teh, editors, \emph{Proceedings of the
  34th International Conference on Machine Learning}, volume~70 of
  \emph{Proceedings of Machine Learning Research}, pages 1607--1616. PMLR,
  06--11 Aug 2017.
\newblock URL \url{https://proceedings.mlr.press/v70/ingraham17a.html}.

\bibitem[Kingma and Welling(2013)]{kingma2013auto}
Diederik~P Kingma and Max Welling.
\newblock Auto-encoding variational bayes.
\newblock \emph{arXiv preprint arXiv:1312.6114}, 2013.

\bibitem[Kipf and Welling(2016)]{kipf2016variational}
Thomas~N Kipf and Max Welling.
\newblock Variational graph auto-encoders.
\newblock \emph{arXiv preprint arXiv:1611.07308}, 2016.

\bibitem[Kipf and Welling(2017)]{Kipf:2017tc}
Thomas~N. Kipf and Max Welling.
\newblock Semi-supervised classification with graph convolutional networks.
\newblock In \emph{International Conference on Learning Representations
  (ICLR)}, 2017.

\bibitem[LeCun et~al.(1995)LeCun, Bengio, et~al.]{lecun1995convolutional}
Yann LeCun, Yoshua Bengio, et~al.
\newblock Convolutional networks for images, speech, and time series.
\newblock \emph{The handbook of brain theory and neural networks},
  3361\penalty0 (10):\penalty0 1995, 1995.

\bibitem[Liao et~al.(2019)Liao, Li, Song, Wang, Hamilton, Duvenaud, Urtasun,
  and Zemel]{gran}
Renjie Liao, Yujia Li, Yang Song, Shenlong Wang, Will Hamilton, David~K
  Duvenaud, Raquel Urtasun, and Richard Zemel.
\newblock Efficient graph generation with graph recurrent attention networks.
\newblock In H.~Wallach, H.~Larochelle, A.~Beygelzimer, F.~d\textquotesingle
  Alch\'{e}-Buc, E.~Fox, and R.~Garnett, editors, \emph{Advances in Neural
  Information Processing Systems}, volume~32. Curran Associates, Inc., 2019.
\newblock URL
  \url{https://proceedings.neurips.cc/paper/2019/file/d0921d442ee91b896ad95059d13df618-Paper.pdf}.

\bibitem[Lin et~al.(2015)Lin, Liu, Sun, Liu, and Zhu]{10.5555/2886521.2886624}
Yankai Lin, Zhiyuan Liu, Maosong Sun, Yang Liu, and Xuan Zhu.
\newblock Learning entity and relation embeddings for knowledge graph
  completion.
\newblock In \emph{Proceedings of the Twenty-Ninth AAAI Conference on
  Artificial Intelligence}, AAAI'15, page 2181–2187. AAAI Press, 2015.
\newblock ISBN 0262511290.

\bibitem[Long et~al.(2022)Long, Pan, Zhang, Son, Kondor, and
  Rzhetsky]{long500molecular}
Yanan Long, Horace Pan, Chao Zhang, Hy~Truong Son, Risi Kondor, and Andrey
  Rzhetsky.
\newblock Molecular fingerprints are a simple yet effective solution to the
  drug--drug interaction problem.
\newblock \emph{The 2022 ICML Workshop on Computational Biology}, 2022.
\newblock URL
  \url{https://icml-compbio.github.io/2022/papers/WCBICML2022_paper_72.pdf}.

\bibitem[Nickel et~al.(2011)Nickel, Tresp, and Kriegel]{nickel2011three}
Maximilian Nickel, Volker Tresp, and Hans-Peter Kriegel.
\newblock A three-way model for collective learning on multi-relational data.
\newblock In \emph{Proceedings of the 28th International Conference on
  International Conference on Machine Learning}, ICML'11, page 809–816,
  Madison, WI, USA, 2011. Omnipress.
\newblock ISBN 9781450306195.

\bibitem[Nyamabo et~al.(2021)Nyamabo, Yu, and Shi]{nyamabo2021ssi}
Arnold~K Nyamabo, Hui Yu, and Jian-Yu Shi.
\newblock Ssi--ddi: substructure--substructure interactions for drug--drug
  interaction prediction.
\newblock \emph{Briefings in Bioinformatics}, 22\penalty0 (6):\penalty0
  bbab133, 2021.

\bibitem[Papalexakis et~al.(2016)Papalexakis, Faloutsos, and
  Sidiropoulos]{papalexakis2016tensors}
Evangelos~E. Papalexakis, Christos Faloutsos, and Nicholas~D. Sidiropoulos.
\newblock Tensors for data mining and data fusion: Models, applications, and
  scalable algorithms.
\newblock \emph{ACM Trans. Intell. Syst. Technol.}, 8\penalty0 (2), oct 2016.
\newblock ISSN 2157-6904.
\newblock \doi{10.1145/2915921}.
\newblock URL \url{https://doi.org/10.1145/2915921}.

\bibitem[Perozzi et~al.(2014)Perozzi, Al-Rfou, and Skiena]{perozzi2014deepwalk}
Bryan Perozzi, Rami Al-Rfou, and Steven Skiena.
\newblock Deepwalk: Online learning of social representations.
\newblock In \emph{Proceedings of the 20th ACM SIGKDD International Conference
  on Knowledge Discovery and Data Mining}, KDD '14, page 701–710, New York,
  NY, USA, 2014. Association for Computing Machinery.
\newblock ISBN 9781450329569.
\newblock \doi{10.1145/2623330.2623732}.
\newblock URL \url{https://doi.org/10.1145/2623330.2623732}.

\bibitem[Ryu et~al.(2018)Ryu, Kim, and Lee]{drug_food}
Jae~Yong Ryu, Hyun~Uk Kim, and Sang~Yup Lee.
\newblock Deep learning improves prediction of drug–drug and drug–food
  interactions.
\newblock \emph{Proceedings of the National Academy of Sciences}, 115\penalty0
  (18):\penalty0 E4304--E4311, 2018.
\newblock \doi{10.1073/pnas.1803294115}.
\newblock URL \url{https://www.pnas.org/doi/abs/10.1073/pnas.1803294115}.

\bibitem[Scarselli et~al.(2009)Scarselli, Gori, Tsoi, Hagenbuchner, and
  Monfardini]{4700287}
Franco Scarselli, Marco Gori, Ah~Chung Tsoi, Markus Hagenbuchner, and Gabriele
  Monfardini.
\newblock The graph neural network model.
\newblock \emph{IEEE Transactions on Neural Networks}, 20\penalty0
  (1):\penalty0 61--80, 2009.
\newblock \doi{10.1109/TNN.2008.2005605}.

\bibitem[Simonovsky and Komodakis(2018)]{graphvae}
Martin Simonovsky and Nikos Komodakis.
\newblock Graphvae: Towards generation of small graphs using variational
  autoencoders, 2018.
\newblock URL \url{https://arxiv.org/abs/1802.03480}.

\bibitem[Suphavilai et~al.(2018)Suphavilai, Bertrand, and
  Nagarajan]{suphavilai2018predicting}
Chayaporn Suphavilai, Denis Bertrand, and Niranjan Nagarajan.
\newblock Predicting cancer drug response using a recommender system.
\newblock \emph{Bioinformatics}, 34\penalty0 (22):\penalty0 3907--3914, 2018.

\bibitem[van~der Maaten and Hinton(2008)]{van2008visualizing}
Laurens van~der Maaten and Geoffrey Hinton.
\newblock Visualizing data using t-sne.
\newblock \emph{Journal of Machine Learning Research}, 9\penalty0
  (86):\penalty0 2579--2605, 2008.
\newblock URL \url{http://jmlr.org/papers/v9/vandermaaten08a.html}.

\bibitem[Veli{\v{c}}kovi{\'{c}} et~al.(2018)Veli{\v{c}}kovi{\'{c}}, Cucurull,
  Casanova, Romero, Li{\`{o}}, and Bengio]{velickovic2018graph}
Petar Veli{\v{c}}kovi{\'{c}}, Guillem Cucurull, Arantxa Casanova, Adriana
  Romero, Pietro Li{\`{o}}, and Yoshua Bengio.
\newblock {Graph Attention Networks}.
\newblock \emph{International Conference on Learning Representations}, 2018.
\newblock URL \url{https://openreview.net/forum?id=rJXMpikCZ}.
\newblock accepted as poster.

\bibitem[Vilar et~al.(2012)Vilar, Harpaz, Uriarte, Santana, Rabadan, and
  Friedman]{vilar2012drug}
Santiago Vilar, Rave Harpaz, Eugenio Uriarte, Lourdes Santana, Raul Rabadan,
  and Carol Friedman.
\newblock Drug—drug interaction through molecular structure similarity
  analysis.
\newblock \emph{Journal of the American Medical Informatics Association},
  19\penalty0 (6):\penalty0 1066--1074, 2012.

\bibitem[Wang et~al.(2020)Wang, Lian, Zhang, Qin, and Lin]{gognn}
Hanchen Wang, Defu Lian, Ying Zhang, Lu~Qin, and Xuemin Lin.
\newblock Gognn: Graph of graphs neural network for predicting structured
  entity interactions.
\newblock In Christian Bessiere, editor, \emph{Proceedings of the Twenty-Ninth
  International Joint Conference on Artificial Intelligence, {IJCAI-20}}, pages
  1317--1323. International Joint Conferences on Artificial Intelligence
  Organization, 7 2020.
\newblock \doi{10.24963/ijcai.2020/183}.
\newblock URL \url{https://doi.org/10.24963/ijcai.2020/183}.
\newblock Main track.

\bibitem[Wang et~al.(2022)Wang, Liu, Shen, Deng, and Liu]{wang2022deepdds}
Jinxian Wang, Xuejun Liu, Siyuan Shen, Lei Deng, and Hui Liu.
\newblock Deepdds: deep graph neural network with attention mechanism to
  predict synergistic drug combinations.
\newblock \emph{Briefings in Bioinformatics}, 23\penalty0 (1):\penalty0
  bbab390, 2022.

\bibitem[Wang et~al.(2017)Wang, Li, Zhang, and Gao]{wang2017improved}
Lin Wang, Xiaozhong Li, Louxin Zhang, and Qiang Gao.
\newblock Improved anticancer drug response prediction in cell lines using
  matrix factorization with similarity regularization.
\newblock \emph{BMC cancer}, 17\penalty0 (1):\penalty0 1--12, 2017.

\bibitem[Wei et~al.(2019)Wei, Liu, Zheng, and Li]{wei2019comprehensive}
Dong Wei, Chuanying Liu, Xiaoqi Zheng, and Yushuang Li.
\newblock Comprehensive anticancer drug response prediction based on a simple
  cell line-drug complex network model.
\newblock \emph{BMC bioinformatics}, 20\penalty0 (1):\penalty0 1--15, 2019.

\bibitem[Wu et~al.(2020)Wu, K\"{o}hler, and Noe]{NEURIPS2020_41d80bfc}
Hao Wu, Jonas K\"{o}hler, and Frank Noe.
\newblock Stochastic normalizing flows.
\newblock In H.~Larochelle, M.~Ranzato, R.~Hadsell, M.F. Balcan, and H.~Lin,
  editors, \emph{Advances in Neural Information Processing Systems}, volume~33,
  pages 5933--5944. Curran Associates, Inc., 2020.
\newblock URL
  \url{https://proceedings.neurips.cc/paper/2020/file/41d80bfc327ef980528426fc810a6d7a-Paper.pdf}.

\bibitem[Xu et~al.(2019)Xu, Wang, Chen, Tao, and Zhao]{mrgnn}
Nuo Xu, Pinghui Wang, Long Chen, Jing Tao, and Junzhou Zhao.
\newblock Mr-gnn: Multi-resolution and dual graph neural network for predicting
  structured entity interactions.
\newblock \emph{Proceedings of IJCAI}, 2019.

\bibitem[Yin et~al.(2022)Yin, Cao, Fan, Liu, Jiang, and Zeng]{yin2022deepdrug}
Qijin Yin, Xusheng Cao, Rui Fan, Qiao Liu, Rui Jiang, and Wanwen Zeng.
\newblock Deepdrug: A general graph-based deep learning framework for drug-drug
  interactions and drug-target interactions prediction.
\newblock \emph{biorxiv}, pages 2020--11, 2022.

\bibitem[Zitnik and Leskovec(2017)]{Zitnik2017}
Marinka Zitnik and Jure Leskovec.
\newblock Predicting multicellular function through multi-layer tissue
  networks.
\newblock \emph{Bioinformatics}, 33\penalty0 (14):\penalty0 190--198, 2017.

\bibitem[Zitnik et~al.(2018)Zitnik, Agrawal, and
  Leskovec]{10.1093/bioinformatics/bty294}
Marinka Zitnik, Monica Agrawal, and Jure Leskovec.
\newblock {Modeling polypharmacy side effects with graph convolutional
  networks}.
\newblock \emph{Bioinformatics}, 34\penalty0 (13):\penalty0 i457--i466, 06
  2018.
\newblock ISSN 1367-4803.
\newblock \doi{10.1093/bioinformatics/bty294}.
\newblock URL \url{https://doi.org/10.1093/bioinformatics/bty294}.

\bibitem[Zong et~al.(2017)Zong, Kim, Ngo, and Harismendy]{zong2017deep}
Nansu Zong, Hyeoneui Kim, Victoria Ngo, and Olivier Harismendy.
\newblock Deep mining heterogeneous networks of biomedical linked data to
  predict novel drug--target associations.
\newblock \emph{Bioinformatics}, 33\penalty0 (15):\penalty0 2337--2344, 2017.

\end{thebibliography}

\end{document}